\PassOptionsToPackage{table}{xcolor}
\documentclass[11pt]{article}
\usepackage{acl}  
\usepackage{times}
\usepackage{latexsym}
\usepackage[T1]{fontenc}
\usepackage[utf8]{inputenc}
\usepackage{microtype}
\usepackage{inconsolata}
\usepackage{graphicx}

\usepackage{algorithm}
\usepackage{bbm}
\usepackage{algpseudocode}
\usepackage{amsmath}
\usepackage{CJK}
\usepackage[table]{xcolor}
\usepackage{booktabs}
\usepackage{todonotes}
\usepackage{tabularx}
\usepackage{longtable}
\usepackage{hyperref}

\title{From BERT to LLMs：Comparing and Understanding \\  Chinese Classifier Prediction in Language Models}

\author{
  \textbf{Ziqi Zhang},
  \textbf{Jianfei Ma}, 
  \textbf{Emmanuele Chersoni},
  \textbf{Jieshun You},
  \textbf{Zhaoxin Feng}\\
   Department of Language Science and Technology, The Hong Kong Polytechnic University \\
  \texttt{\{blameredens.zhang, jianfei-mark.ma,} \\
  \texttt{jieshun.you, zhaoxinbetty.feng\}@connect.polyu.hk,} \\
  \texttt{emmanuele.chersoni@polyu.hk}
}

\begin{document}
\begin{CJK}{UTF8}{gkai}
\maketitle
\begin{abstract}
Classifiers are an important and defining feature of the Chinese language, and their correct prediction is key to numerous educational applications. Yet, whether the most popular Large Language Models (LLMs) possess proper knowledge the Chinese classifiers is an issue that has largely remain unexplored in the Natural Language Processing (NLP) literature.

To address such a question, we employ various masking strategies to evaluate the LLMs' intrinsic ability, the contribution of different sentence elements, and the working of the attention mechanisms during prediction. Besides, we explore fine-tuning for LLMs to enhance the classifier performance.

Our findings reveal that LLMs perform worse than BERT, even with fine-tuning. The prediction, as expected, greatly benefits from the information about the following noun, which also explains the advantage of models with a bidirectional attention mechanism such as BERT. 
\end{abstract}

\begin{figure}[t]
  \includegraphics[width=0.48\textwidth]
  {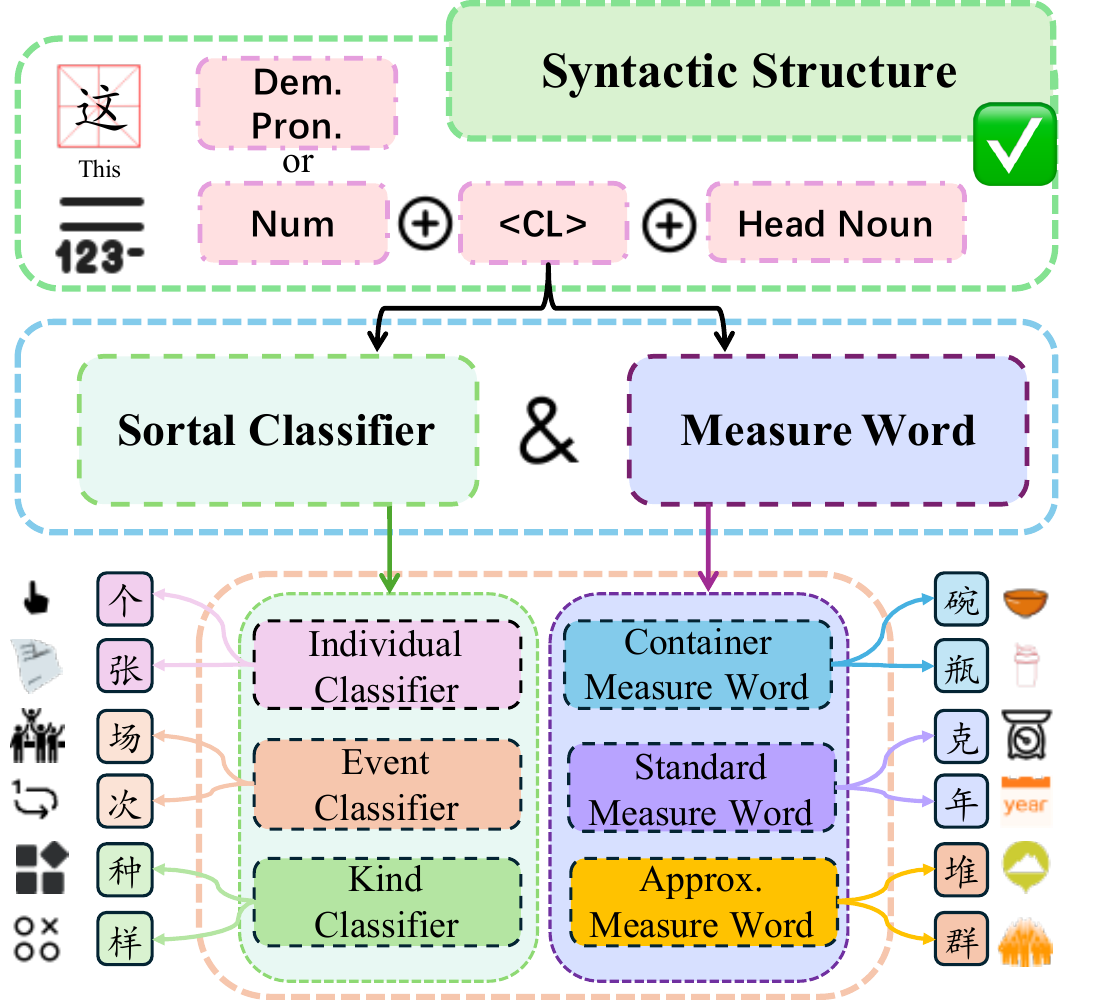} 
  \caption{This figure elaborates on correct syntactic structures of the Chinese classifier system, the types of Chinese classifiers, and corresponding examples, where Num, CL, Dem. Pron., and Approx. stand for numeral, classifier, demonstrative pronoun, and approximation, respectively. Notably, several classifier examples on both sides are accompanied by icons that illustrate the approximate meanings they convey. A detailed explanation is provided in Appendix~\ref{ref: Chinese Classifier Categories and Explanation}.}
  \label{fig:fig1}
\end{figure}

\section{Introduction}
Chinese classifiers constitute a morphosyntactic category that semantically marks noun classes~\cite{intro_classifier_detail}. They precede a head noun and combine with numerals or demonstrative pronouns to convey quantity or frequency within noun phrases~\cite{Li1989}. As illustrated in Figure~\ref {fig:fig1}, such linguistic devices construct a complex system describing different semantic features of head nouns that they precede~\citep{Huang2016-gm}. The large classifier inventory in Chinese often allows different classifiers to combine with the same head noun, conveying distinct semantic nuances \citep{Shi2014-wi,Huang2014-zt}. For example, both individual classifiers ``个'' and ``位'' can modify the noun of people, while the former is a more generic one, the latter is restricted to highly-regarded professions and conveys a polite tone. Improper collocations can result in semantic or pragmatic violations~\cite{Chan_2019}.

In the natural language processing (NLP) literature, there has been extensive exploration of Natural Language Understanding (NLU) using Pre-trained Language Models (PLMs)~\cite{Wang2022-tx} and Large Language Models (LLMs)~\cite{10.1145/3641289,ma-etal-2025-reasoning}. However, all current studies on Chinese classifiers only include behaviours of traditional models for classifier prediction or selection with limited interpretability efforts ~\cite{peinelt-etal-2017-classifierguesser,jarnfors-etal-2021-using}, while to the best of our knowledge, there is no evaluation study with more recent autoregressive LLMs. In addition to the interest in evaluating LLMs on this essential component of Chinese grammar, it should be kept in mind that learning classifier systems has been proven to be particularly challenging for learners of Chinese as L2 \cite{liang2008acquisition, liu2018l2}, and thus NLP technologies with a robust knowledge of classifiers would be a precious resource to develop educational tools.



Given the above-mentioned gap in the literature, we address the following research questions: How do LLMs perform in this classifier prediction? What are the semantic contributions of the different elements of a sentence to the process of selecting a classifier, and can this be observed from the attention mechanism of the model?

With this goal in mind, we establish a control task by iteratively inserting various classifiers into a blank classifier position within sentence and ranking them based on their Language Model (LM) log probabilities. With the same setups, we randomly extract sentence samples and mask the token of the classifier and then fine-tune the models to examine how well they can perform. Finally, we carry out additional analysis by modifying the model's attention mask, in order to make them ignore the surrounding words in the sentence and quantify their information contribution to task performance. 

The control task shows that BERT~\cite{devlin-etal-2019-bert} and LLMs achieve good accuracy in the prediction, yet the former exhibits a distinct advantage and higher improvement potential with fine-tuning. Due to the strong semantic link between classifiers and head nouns, models exhibit a high dependency on the corresponding head noun during prediction. Intriguingly, the results also confirm an additional (albeit weak) contribution from the remaining contextual information. The same experiment also reveals that the bidirectional attention mechanism plays a critical role, despite the bigger parameters and training data size of autoregressive LLMs.



\section{Related Work}
\subsection{Chinese Classifiers}

Chinese classifiers serve as obligatory syntactic elements bridging numerals and head nouns, forming grammatically complete noun phrases \citep{Li1989} while encoding semantic features, including shape and function, and taxonomic categorization \citep{Lakoff1986-gp,CroFT1994SemanticUI}. Current research on this element centers on the usage patterns across diverse population groups~\cite{zhan-levy-2018-comparing,shi-2021-schizophrenia} and its nuanced idiosyncrasies~\cite{liu-etal-2019-idiosyncrasies}. 

However, recent computational studies of the prediction task remain scarce. Existing studies most focus on early approaches, covering SVMs \citep{guo-zhong-2005-chinese} and Word2Vec embeddings \citep{peinelt-etal-2017-classifierguesser}, later augmented with mutual information metrics \citep{liu-etal-2019-idiosyncrasies}.

The Transformers marked a turning point. But only \citet{jarnfors-etal-2021-using} demonstrated BERT's superior performance after fine-tuning, though revealing persistent deficiencies in implication covering politeness and plural markers. This limitation motivates investigating whether modern LLMs' enhanced contextual awareness and linguistic knowledge can achieve more robust classifier prediction.

\subsection{Attention Mechanism in Lexical Semantics}

BERT's bidirectional attention provides comprehensive contextual awareness by processing both left and right contexts of target words~\cite{devlin-etal-2019-bert}. This architectural advantage has been applied to LLMs and empirically validated across NLP tasks, like syntactic parsing and named entity recognition \cite{behnamghader2024llm2vec,springer2025repetition}. Building on this foundation, \citep{feng2025learning} demonstrates that bidirectional architectures particularly excel in semantic tasks requiring precise context resolution with the framework constructed by~\citet{behnamghader2024llm2vec}.


While autoregressive LLMs are inherently constrained by unidirectional attention, their substantially expanded pretraining corpora and enhanced world knowledge \citep{wei2022emergent,llmasfsl} might offer compensatory advantages across various NLU tasks. However, specifically, for Chinese classifier prediction, the trade-off effect of this architectural dichotomy on classifiers remains unexamined. Furthermore, how bidirectional attention boosts BERT's accuracy and how head nouns impact the performance of bidirectional models both require investigation.




\subsection{Masking Strategies for Probing}

Masking strategies enable controlled experiments by selectively processing target regions to assess performance changes or predicted output~\cite{petroni-etal-2019-language,kassner-schutze-2020-negated,zhong-etal-2021-factual}. One of the typical approaches is to modify LMs' attention masks, zeroing selected token weights to study attention mechanisms' effects~\cite{liong-etal-2024-unveiling}. In the computational linguistics domain, this approach helps assess specific linguistic components' contributions. For instance, \citet{metheniti-etal-2020-relevant} showed that masking non-verbal linguistic elements improves BERT's alignment with human intuitions for role fillers, while \citet{cho-etal-2021-modeling} demonstrated similar benefits for event location prediction by masking context and forcing attention on verb phrases.

Following the previous work, we mask the target classifier to trigger the model's prediction at this position and adjust the attention mask to examine how effectively context-based attention contributes to the target classifier.

\subsection{Classifier Ranking by Log Probability}

Log Probability (LogProb) has been proven effective for various token-level tasks like assessing grammatical correctness and semantic plausibility, where its outputs often align with human judgments and outperform direct prompting~\cite{hu2023prompting,kauf-etal-2024-log}.

We acknowledge that this metric is not without limitations. It is known to be sensitive to confounding variables such as word frequency\footnote{The analysis of effect for word frequency on accuracy is attached in Appendix~\ref{ref: Relationship of classifier frequency and accuracy}} and output length~\cite{salazar-etal-2020-masked,holtzman-etal-2021-surface}. But they are difficult to isolate due to the uneven distribution nature of pre-training data. Despite this sensitivity, LogProb also be recognized as the robust choice for check the certainty of LMs' output because its validity and superiority for semantic tasks are strongly supported by~\citet{kauf-etal-2024-log}.

To specifically mitigate the influence of output length,we insert classifier candidates into a sentence to compute the average LogProb of each filled sentence to directly obtain the score without any redundant generation  In this framework, the candidate that yields the highest sentence-level LogProb is considered the best fit.

\section{Methodology}

\begin{figure*}[h]
\centering
\includegraphics[width=\textwidth]{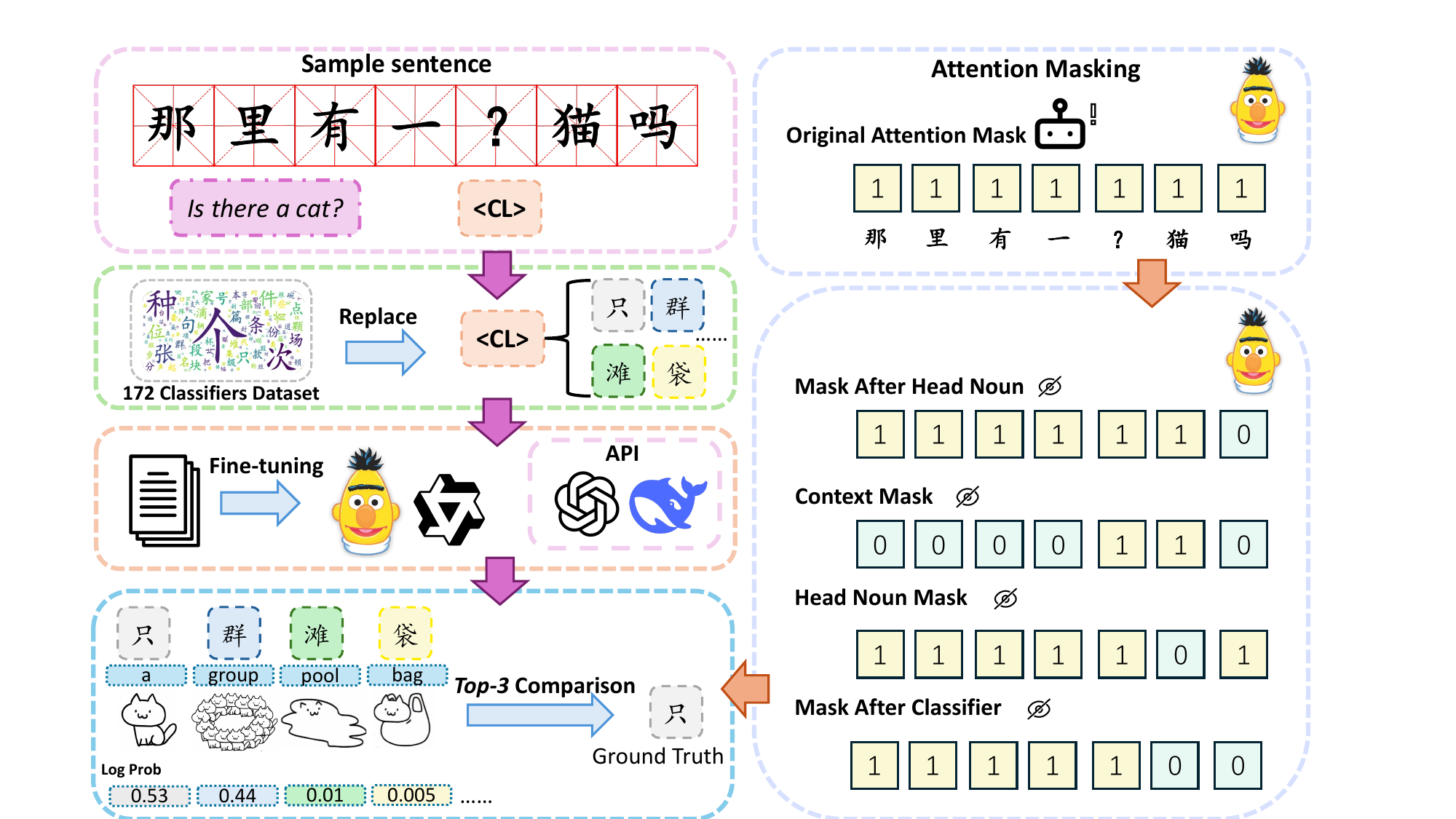}
\caption{This figure shows the workflow of the project. As denoted by purple arrows, the given sample sentence is input into LMs for prediction. The sentences in the dataset vary in length (and are not necessarily limited to 7 tokens as in the example shown in the figure), while classifiers may consist of either one or two Chinese characters.} \label{fig3}
\end{figure*}

This study evaluates the performance of two types of LMs in Chinese classifier prediction: (1) the model with bidirectional attention mechanisms, BERT, with masked language modeling and fine-tuning; and (2) autoregressive LLMs, including local deployments (Qwen3-1.7B, 4B, 8B and corresponding fine-tuned versions) and full-parameter APIs (DeepSeek-R1 and GPT-4). Due to the complex mapping between head nouns and classifiers, we obtain accuracy based on log probability ranking for evaluation. The detailed workflow is demonstrated in Figure~\ref{fig3}.


\subsection{Dataset Constructions}
We employ the Chinese Classifier Dataset \cite{peinelt-etal-2017-classifierguesser}, a comprehensive resource with annotated classifier-noun pairs in sentential contexts, convenient to adapt classifier prediction tasks. This dataset contains 681,104 sentences, encompassing 172 distinct classifiers that nearly cover the entire commonly used Mandarin classifiers. Additionally, the Stanford constituent parser~\cite{levy-manning-2003-harder} was applied to annotate the head noun in each sentence. Although classifiers exhibit diverse pairings in pragmatic contexts, their syntactic component combinations are highly fixed, as supported by ~\cite{Li1989}, suggesting that a relatively small number of examples is sufficient to effectively evaluate their accurate usage and prediction. Due to this and computational resource limitations, we initially randomly sampled 11,986 sentences that span all classifiers and preserve their original distribution. After manual screening to remove 69 erroneous cases (e.g., annotation errors or syntactic anomalies), we obtained 11,917 valid sentences for further processing. These sampled instances were split into training and test sets at an 85:15 ratio to support fine-tune and evaluation. 


\subsection{BERT Classifier Prediction}
\textbf{Masked language modeling} \quad To evaluate BERT's performance in Chinese classifier prediction, we utilize the \textbf{BERT-base-chinese} model through masked language modeling (MLM). Our approach computes the conditional probability of candidate classifiers at the masked position, accommodating both single-token classifiers (e.g., ``个'') and two-token classifiers (e.g., ``档子''). Given a tagged sentence \( X = (x_1, \ldots, \text{<CL>}, \text{<h>}, \text{head noun}, \text{</h>}, \ldots, x_n) \), where \text{<CL>} is the placeholder for the classifier and \text{<h>}, \text{</h>} demarcate the head noun, we replace \text{<CL>} with one or two ``[MASK]'' tokens based on the classifier's tokenization. We calculate the log probability for each candidate classifier \( c \in C \), where \( C \) is a set of 172 classifiers, encompassing both single-character and two-character classifiers.

For single-token classifiers, the log probability of a classifier \( c \) is computed as:
\begin{equation}
\scalebox{0.9}{$
\log P(c | X) = \log \left( \text{softmax} \left( \text{BERT}(X_f(c))_{[I_1]} \right)_c \right)
$}
\end{equation}
where \( X_f(c) \) is the sentence with \text{<CL>} replaced by classifier \( c \), \( I_1 \) denotes the position of the single "[MASK]" token, and \( \text{softmax}(\cdot)_c \) represents the probability of classifier \( c \).

For two-token classifiers, where \( c = (c_1, c_2) \), the joint log probability is calculated as:
\begin{equation}
\scalebox{0.8}{$
\log P(c | X) = \sum_{m=1}^{2} \log \left( \text{softmax} \left( \text{BERT}(X_f(c))_{[I_m]} \right)_{c_m} \right)
$}
\end{equation}
where \( I_m \) is the position of the \( m \)-th \texttt{[MASK]} token (\( m = 1, 2 \)), and \( \text{softmax}(\cdot)_{c_m} \) is the probability of the \( m \)-th token of classifier \( c \). The joint log probability sums the log probabilities of both mask positions, accurately capturing the combined likelihood of the two tokens.

\noindent \textbf{Fine-tuning} \quad We use the full training set over 3 epochs with the AdamW optimizer (learning rate: $2 \times 10^{-5}$) with early stopping strategy. 

\subsection{LLM-Based Classifier Prediction}

\textbf{Sentence log probability} \quad Unlike BERT, we utilize sentence-level log probabilities for classifier ranking due to the autoregressive nature of LLMs. Since they can only access leftward context when predicting the classifier token, the isolated token probability fails to incorporate crucial information about the subsequent noun or other sentence elements. This lack of right-context access renders token-level probabilities unreliable for our task.

With locally deployed Qwen3, we replace the empty classifier position (indicated with an underscore) in each sentence with each of 172 candidate classifiers and use the \textbf{IncrementalLMScorer} from the \textbf{minicons}\footnote{\textbf{minicons} is a Python library for efficient probability scoring of transformer-based language models. Please refer to its' github link: https://github.com/kanishkamisra/minicons} to extract the log probability of each filled sentence by averaging the token-level log probabilities.It can be represented as:

\begin{equation}
\log P(S_c | X) = \frac{1}{T} \sum_{t=1}^{T} \log \left( P(w_t | w_{<t}, X_f(c)) \right)
\end{equation}
where \( S_c \) is the sentence with classifier \( c \) inserted, \( X_f(c) \) is the sentence with \text{<CL>} replaced by \( c \), \( T \) is the total number of tokens, \( w_t \) is the \( t \)-th token, and \( w_{<t} \) is the preceding context. This approach evaluates the overall coherence of the sentence with the inserted classifier, averaging the log probabilities of all tokens to normalize for sentence length.


\noindent \textbf{Prompting via API} \quad For the full-parameter models DeepSeek-R1 and GPT-4, we designed prompts to guide them to generate the most probable Chinese classifier for each given sentence with an empty classifier position. To diminish extraneous reasoning and maintain diversity of responses, we configured the temperature to 0, top-p to 0.9, and maximum token length to 32. To ensure uniqueness, outputs are further refined using a set-based deduplication method. 


For GPT-4, we set \textbf{logprobs} parameter to be true in the API call, enabling the model to return the logarithmic probabilities of each output token. Thus, we can ensure that predicted classifiers can be sorted by their log-probability in descending order as Qwen. DeepSeek-R1 API, however, does not support LogProb extraction. Hence, we perform repeated generations with multiple candidate outputs and select the first result containing three distinct single-character classifiers as formal selections. 


\noindent \textbf{Metrics} \quad Predictions were evaluated using two metrics: \textit{Accuracy} and \textit{R-Rank}. \textit{Accuracy} measures the proportion of samples where the model's top predicted classifier matches the correct classifier. \textit{R-Rank}, based on previous work \cite{camacho2018semeval,peng2022discovering}, evaluates the model's nuanced understanding of classifier selection by considering the rank of the correct classifier within the top 3 predictions. Specifically, for each sample \(i\), we define \(rank_i\) as the rank of the correct classifier among the top 3 predictions, or 4 if it is not among them. These metrics are defined as follows:

\begin{equation}
\text{Accuracy} = \frac{1}{n} \sum_{i=1}^{n} \mathbbm{1}\left( y_i = y_i^l[0] \right)
\end{equation}
where \(n\) is the total number of samples, \(y_i\) is the correct classifier for the \(i\)-th sample, \(y_i^l[0]\) is the model's top predicted classifier for the \(i\)-th sample, and \(\mathbbm{1}\left( y_i = y_i^l[0] \right)\) is an indicator function that returns 1 if the top prediction matches the correct classifier, and 0 otherwise.

\begin{equation}
\text{R-rank} = \frac{1}{n} \sum_{i=1}^n rank_i
\end{equation}
where \(n\) is the total number of samples, and \(rank_i\) is the rank of the correct classifier for the \(i\)-th sample within the model's top 3 predictions (1, 2, or 3), or 4 if it is not in the top 3. 
The choice of a top-3 cutoff for R-rank was deliberate. Given that many Chinese nouns collocate with multiple classifiers, evaluating the top-3 candidates provides a sufficiently broad and practical assessment of a model's discriminative ability. Extending this range further would dilute the metric's practical significance, as lower-ranked candidates are increasingly unlikely to be contextually appropriate.

\section{Experimental Results and Analyses}

\subsection{Can LLMs Be Good Classifier Guessers?}

\begin{table}[htbp]
\centering
\resizebox{\columnwidth}{!}{
\begin{tabular}{lcc}
\toprule
\textbf{Model} & \textbf{Accuracy} & \textbf{R-rank} \\
\midrule
BERT MLM  & 62.31 & 1.8298 \\
BERT-ft & \textbf{69.54} & \textbf{1.6676} \\
GPT-4 & 50.70 & 2.1408 \\
DeepSeek-R1 & 59.64 & 1.9400 \\
Qwen3-1.7B & 31.80 & 2.7821 \\
Qwen3-1.7B-ft & 39.03 & 2.5107 \\
Qwen3-4B & 33.46 & 2.7270 \\
Qwen3-4B-ft & 47.69 & 2.2698 \\
Qwen3-8B & 39.03 & 2.5107 \\
Qwen3-8B-ft & 39.94 & 2.4861 \\
\bottomrule
\end{tabular}
}
\caption{Results of \textit{accuracy} and \textit{R-rank} of different LMs for Chinese classifier prediction. ``MLM'' stands for ``Masked Language Modeling'', and ``ft'' denotes ``Fine-tuning''.}
\label{tab:model_performance}
\end{table}

The results in Table~\ref{tab:model_performance} demonstrate BERT's superior performance, achieving both the highest \textit{accuracy} and the optimal \textit{R-rank} scores, suggesting its effectiveness in Chinese classifier prediction. In contrast, autoregressive LLMs, including GPT-4 and the Qwen3 variants, generally underperform, with most models failing to surpass 0.5 \textit{accuracy} and exhibiting \textit{R-rank} values between 2 and 3. Notably, Deepseek-R1 is an exception, achieving a competitive R-rank and higher \textit{accuracy} than other LLMs, though it still falls short of BERT's performance. While scaling model parameters yields marginal improvements, even the largest models in this study, Deepseek-R1 and GPT-4, do not close the gap with BERT. This suggests that architectural differences (e.g., masked language modeling vs. autoregressive generation) may play a more critical role than parameter size in this task.

\subsection{Can LLMs with Fine-tuning Close the Performance Gap to BERT?}
Different sizes of Qwen3 models exhibit significant improvements in both \textit{accuracy} and \textit{R-rank} after fine-tuning in Table~\ref{tab:model_performance}. Interestingly, the scaling effect of model parameters does not align with the default models' performance trends, and the Qwen3-4B-ft achieves optimal performance in both metrics among the Qwen3 variants.

However, while fine-tuning leads to substantial performance gains, the enhanced Qwen3-4B model only reaches \textit{accuracy} levels comparable to GPT-4, still falling significantly short of BERT's performance. Furthermore, when applying the same fine-tuning procedure to BERT, we observe an inverse relationship between the two metrics. Despite this, the fine-tuned LLMs fail to match BERT's performance in either metric, suggesting that fine-tuning alone may be insufficient to overcome LLMs' inherent limitations in classifier prediction tasks.

\subsection{Can LLMs Balance Prediction Performance among Classifier Types?}

\begin{figure*}[h]
\centering
\includegraphics[width=\textwidth]{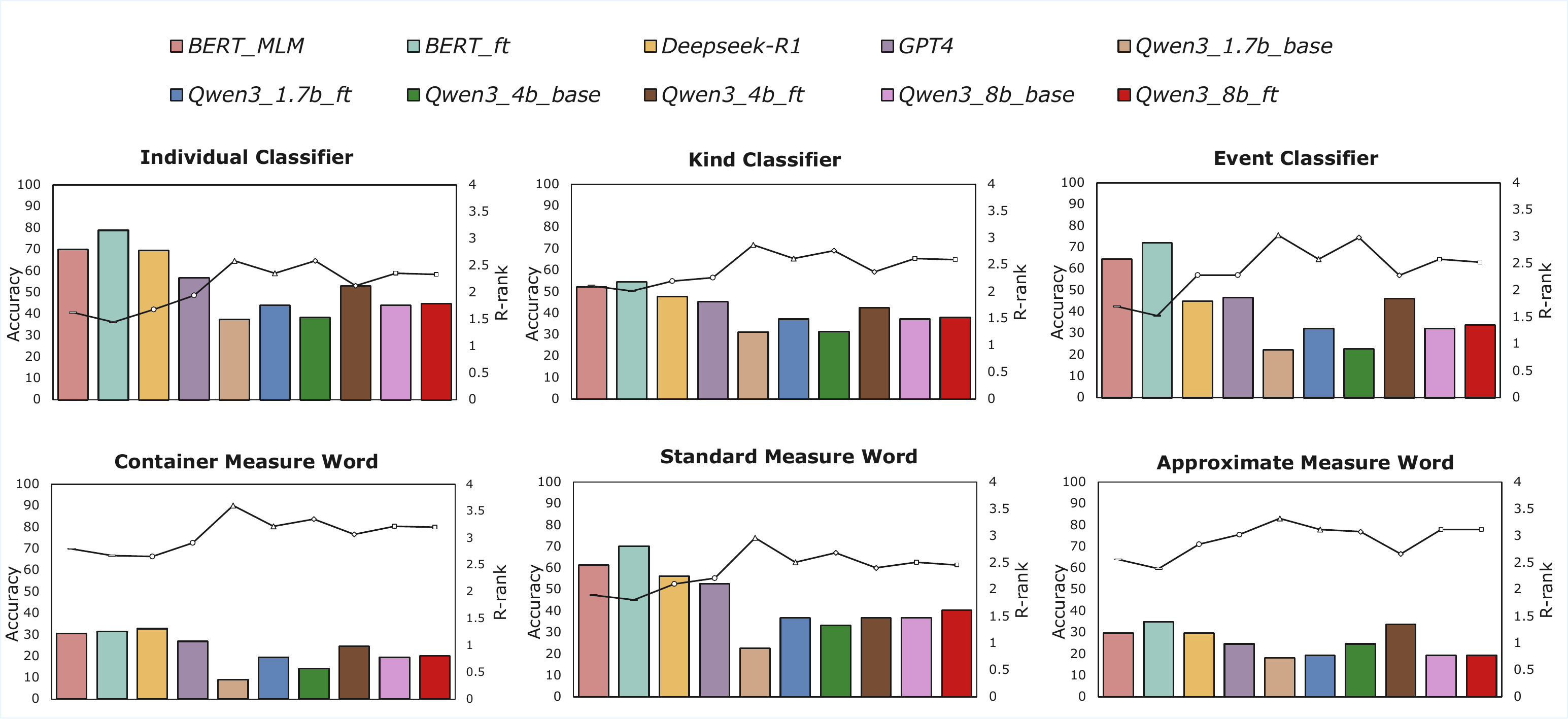}

\caption{The fine-grained analysis of the six types of classifiers' predictions among the proposed LMs. The black polyline represents the \textit{R-rank} value (the lower, the better); the bar charts in different colors represent the \textit{accuracy} of specific models in this type of classifier (the higher, the better). The ``ft'' in model names represents that the LMs have applied a fine-tuning strategy.} 
\label{fig:fig3}
\end{figure*}

While LLMs currently trail BERT in overall performance, their potential to leverage vast pre-training data to address BERT's key limitations, particularly inconsistent performance across task types and weaker fine-grained semantic discrimination, warrants further investigation. This motivates our detailed analysis of classifier performance across different task types and models.

As illustrated in Figure~\ref{fig:fig3}, we evaluate models' \textit{accuracy} per classifier type (\textit{R-rank} with similar trends). Contrary to expectations, LLMs fail to perform more balanced or superior semantic precision and R-rank than BERT despite their broader pretraining; in many cases, they lag behind.

For sortal classifiers, the individual classifiers yield the highest \textit{accuracy} across models, likely due to their reliance on explicit head-noun features, straightforward classification logic, and high frequency in training data. However, event classifiers reveal only a marginal gap between BERT and LLMs, suggesting comparable challenges in modeling event semantics for both of them. Notably, BERT's strong performance in kind classifiers, paired with LLMs' decline, highlights the latter's typological understanding deficits.

The performance of LMs' measure classifiers reveals an important distinction, while standard measure classifiers achieve relatively strong results across all models due to their rigid syntactic patterns, both BERT and LLMs struggle with container and approximate classifiers. This performance dichotomy suggests that while models can effectively learn predictable, formulaic relationships, they face fundamental challenges in modeling more complex items like the container-contents relationship and quantifying abstract concepts.

The similar performance patterns between LLMs and BERT, coupled with LLMs' overall weaker results, suggest that neither their expanded pre-training data scale nor their enhanced capabilities from larger parameters lead to improved prediction performance. This persistent performance gap may warrant further investigation into architectural differences, particularly in attention mechanisms.

\subsection{How LMs' Attention Mechanisms Contribute to Prediction?}

\begin{table}[htbp]
\centering
\fontsize{10pt}{4pt}
\renewcommand{\arraystretch}{1.2} 
\begin{tabular}{@{}m{3.5cm}>{\raggedright\arraybackslash}m{3.5cm}@{}}
\toprule
\textbf{Attention Mask Type} & \textbf{Token Visibility Pattern} \\
\midrule
Standard & {[CLS] 那 里 有 一 [MASK] 猫 吗 ? [SEP]} \\
\addlinespace[2pt]
Mask After Head Noun & {[CLS] 那 里 有 一 [MASK] 猫 0 0 [SEP]} \\
\addlinespace[2pt]
Context Mask & {[CLS] 0 0 0 0 [MASK] 猫 吗 ? [SEP]}  \\
\addlinespace[2pt]
Head Noun Mask & {[CLS] 那 里 有 一 [MASK] 0 吗 ? [SEP]} \\
\addlinespace[2pt]
Mask After Classifier & {[CLS] 那 里 有 一 [MASK] 0 0 0 [SEP]} \\
\bottomrule
\end{tabular}
\caption{Token visibility patterns under different masking strategy types. The positions with 0s correspond to tokens in the input for which the model's attention is blocked(instead of 0 in the text sequence). }
\label{tab:attention_mask_types}
\end{table}

Given the classifiers' strong dependence on their head nouns, the differences in attention mechanisms between BERT and LLMs, and the above analysis, we further investigate how the architectural distinctions account for the gaps. We select BERT as our baseline reference and employ 4 different attention masking types distinct from the standard attention masking for BERT MLM. Examples and comparisons are shown in Figure~\ref{tab:attention_masking}.

Inspired by \citet{metheniti-etal-2020-relevant}, we design four masking strategies by zeroing out tokens in BERT's attention mask, shown in Table~\ref{tab:attention_mask_types}.



The results of adjusting attention masking in Table \ref{tab:attention_masking} show an obvious decline trend in both \textit{accuracy} and \textit{R-rank}. The minor decrease in performance for Mask After Head Noun and Context Mask indicates that directional contextual information (excluding the head noun) contributes marginally to prediction. Based on the changes in both metrics, the text after the head noun has a greater influence on R-rank, while the preceding content affects \textit{accuracy} more significantly.

\begin{table}[htbp]
\centering
\label{tab:accuracy_metrics}
\begin{tabular}{lcc}
\toprule
\textbf{Attention Mask Type} & \textbf{Accuracy}& \textbf{R-rank} \\
\midrule
Standard &  \textbf{62.31} & \textbf{1.8298}  \\
Mask After Head Noun & 60.92 & 1.8929 \\
Context Mask & 58.35 & 1.9272 \\
Head Noun Mask & 33.19 & 2.6670 \\
Mask After Classifier & 25.59& 2.9443 \\
\bottomrule
\end{tabular}
\caption{Performance for BERT with various attention masking strategies.}
\label{tab:attention_masking} 
\end{table}

When the head noun is masked, the performance plummets compared to the standard condition, highlighting the high dependency of classifier prediction on language models. However, further masking the preceding context reveals an interesting pattern where \textit{accuracy} experiences a significant drop, \textit{R-rank} performance shows a slight rebound.  This scenario mirrors LLM's unidirectional attention mechanism. With the scale of 110M parameters, BERT achieves only around 25\% \textit{accuracy} and a ranking score near 3. This result underscores the critical role of the bidirectional attention mechanism, which doubles the \textit{accuracy} while reducing the ranking score by one.

The strong dependency on the head noun and the partial dependence on preceding contexts in classifier prediction seem to strictly require bi-directional attention for efficient modeling. This explains why increasing the parameter and training data size fail to compensate for the inherent limitations of the attentional mechanism.

\section{Error Case Analysis}

Although current LMs, particularly BERT, exhibit strong capabilities in Chinese classifier prediction, their varying performance across different models and various classifier categories underscores persistent challenges. To better understand these error patterns, we systematically analyze two primary types of failures, aiming to empirically investigate the underlying causes of these specific errors.

\subsection{Unable to Capture Fine-grained Pragmatic Preferences}

\begin{table}[htbp]
\centering
\label{tab:accuracy_metrics}
\begin{tabular}{lp{1.75cm}p{3.5cm}}
\toprule
\textbf{Model} & \textbf{Predictions} & \textbf{Carrier Sentence} \\
\midrule
BERT & 件, 种, 回 & 早起这\textcolor{red}{件}事是要多痛苦有多痛苦。 \\
Qwen3 & 件, 个, 桩 & 早起这\textcolor{red}{件}事是要多痛苦有多痛苦。 \\
GPT-4 & 件, 桩, 回 & 早起这\textcolor{red}{件}事是要多痛苦有多痛苦。 \\
Ds-r1 & 件, 桩, 种 & 早起这\textcolor{red}{件}事是要多痛苦有多痛苦。 \\
\bottomrule
\end{tabular}
\caption{The models' responses demonstrate the failure to capture fine-grained pragmatic preferences. The most appropriate candidate is ``档子''. BERT and Qwen3 results are selected from the top three results of the base and fine-tuning models with the best performance. Ds-r1 denotes Deepseek-R1. The English translation of the carrier sentence is \textbf{``Getting up early, this thing is as painful as it get''}.}
\label{tab:error1} 
\end{table}

Current language models demonstrate systematic shortcomings in aligning with pragmatic preferences when selecting classifiers, consistently favoring statistically frequent but stylistically inappropriate options. As illustrated in Table~\ref{tab:error1}, the models' universal top prediction of ``件'' (piece) in a colloquial negative-affect context, followed by other generic or semantically mismatched classifiers like ``种'' (kind) and ``回'' (occasion), reveals their inability to integrate register, affective tone, and habitual semantics into classifier choice. While all models recognize grammatical validity, they diverge in subsequent errors. Qwen3 persists with generic classifiers, full-parameter LLMs incorrectly shift toward event classifiers, and BERT shows partial awareness of categorical distinctions, yet all share the critical failure to prioritize the pragmatically optimal ``档子'', which uniquely satisfies colloquialism, negative affect, and abstract habitual semantics.

This consistent neglect of stylistic and affective dimensions underscores that LMs treat classifier selection as a frequency-driven grammatical task rather than a pragmatic negotiation between linguistic constraints and communicative intent. The hierarchy of error from grammatical correctness to semantic coherence to pragmatic appropriateness exposes their inability to progress beyond coarse statistical patterns toward fine-grained sociolinguistic competence.

\subsection{Hardly Further Check Whole Context}

\begin{table}[htbp]
\centering
\small
\label{tab:accuracy_metrics}
\begin{tabular}{lp{1.75cm}p{3.5cm}}
\toprule
\textbf{Model} & \textbf{Predictions} & \textbf{Carrier Sentence} \\
\midrule
BERT & 笔, 支, 把 & 后来抽奖,又抽到一\textcolor{red}{笔}笔, 虽不算好,总比什么都没有的人强。 \\
Qwen3 & 支, 把, 枝 & 后来抽奖,又抽到一\textcolor{red}{支}笔,虽不算好,总比什么都没有的人强。 \\
GPT-4 & 支, 件, 份 & 后来抽奖,又抽到一\textcolor{red}{支}笔, 虽不算好,总比什么都没有的人强。 \\
Ds-r1 & 支, 管, 杆 & 后来抽奖,又抽到一\textcolor{red}{支}笔, 虽不算好,总比什么都没有的人强。 \\
\bottomrule
\end{tabular}
\caption{With similar descriptions and settings as Table~\ref{tab:error1}. This table demonstrates the LMs may not check all the context for classifier selection. The English translation of the carrier sentence is \textbf{``Later in the raffle, I drew one pen, not great, but better than nothing''}. The proper classifier is ``盒''(box). The classifiers' meanings are``支''(stick),``把''(grasp),``枝''(branch),``件''(piece),`` 份''(portion),``管''(pipe), and ``杆''(rod).}
\label{tab:error2} 
\end{table}

Current language models demonstrate a concerning tendency to make classifier predictions based on local noun-classifier associations rather than holistic context evaluation. This limitation becomes evident when examining BERT's performance in the raffle scenario, where its top prediction ``笔'' (pen) reveals a fundamental misunderstanding. While ``一笔笔'' could theoretically form a plural classifier for money, this interpretation completely disregards the actual context of awarding pens as a prize. Its subsequent predictions, though grammatically correct for describing individual pens, still fail to account for the pragmatic implausibility of awarding just one pen in a raffle setting, a scenario that typically involves more prizes unless explicitly stated otherwise.

The comparative performance of Deepseek-R1, which generated BERTs' subsequent similar candidates. While these properly match the semantic requirements for slender objects, they also overlook the unlikelihood of the single-pen raffle scenario. More alarmingly, Qwen and GPT exhibit even more severe limitations, producing completely unacceptable classifier-noun combinations in their secondary predictions. This degradation in performance highlights how advanced LLMs frequently fail to progress beyond basic noun-classifier matching to consider broader context.

While all models demonstrate basic grammatical competence in noun-classifier pairing, their ability to incorporate pragmatic considerations varies significantly. The most sophisticated models (like Deepseek-R1) at least maintain grammatical accuracy, whereas others (particularly Qwen and GPT) degrade to producing outright errors when forced beyond their primary predictions. This indicates current LMs lack robust mechanisms for contextual integration, instead relying on progressively weaker fallback strategies when their initial predictions prove contextually inadequate. The models' consistent failure to question the plausibility of the single-pen raffle scenario particularly illustrates their limited capacity for real-world reasoning.

\section{Conclusions}
Our study compares the performance of BERT and LLMs in Chinese classifier prediction, revealing that LLMs still underperform compared to BERT, and highlighting the critical role of attention mechanisms. While advanced LLMs possess strengths such as rich world knowledge and fine-grained semantic sensitivity, our results prove that BERT, with or without fine-tuning, achieves better performance in the task. Strikingly, when preceding attention is masked, BERT's performance declines sharply, even falling below that of Qwen3-1.4B.

This explains why LLMs with extensive knowledge bases still demonstrate a significant performance gap even when using enhanced prompts or fine-tuning. The inherent limitation lies in their unidirectional attention mechanism, which fundamentally constrains their effectiveness for this task. These findings highlight the critical role of bidirectional attention and suggest that future research should focus on new strategies to enhable bidirectional attention in LLMs, in order to combine the strengths of both architectures and advance Chinese classifier prediction performance.




\section*{Ethics Statement}
We do not foresee any ethical risks related to our research.

\section*{Limitations}
Though this research provides insights on the comparative performance of BERT and LLMs for Chinese classifier prediction, there are also limitations that should be acknowledged.

First, the evaluation methodology of BERT and LLMs has to be different due to architectural formula differences. Specifically, BERT, as an Encoder,  allows single-token log probability retrieval for tokens that have been masked, but Decoder model like DeepSeek cannot produce log probability for a single token and only provide sentence-level average log probability instead. Furthermore, there are LLM APIs that do not even provide token-level log probabilities, thereby inevitably adding dissimilarities in the model being assessed's performance, across architectures.

Second, the log probability measurements utilized in our model forms have been shown to be a function of sentence length and word frequencies. Sentence lengths in this piece were not held fixed, and thus confounding factors may have been introduced as a consequence. Accordingly, differences seen across sentences and models cannot be separated fully from variations in sentence format, with the possible consequence of reducing the objectivity and interpretability of the findings.

Third, the evaluation dataset includes annotation ambiguities, especially in identifying the head noun. Imperfect or inconsistent annotation may corrupt the training as well as the evaluation performance, introducing noise and potential bias into the reported results.

Finally, this study does not explore the full spectrum of fine-grained semantic distinctions present within the Chinese classifier system. Subtle nuances between classifier usage. For example, those dependent on pragmatic or context-specific cues remain under-investigated and could represent important directions for further analysis.

Future work may address these limitations by standardizing evaluation metrics, curating high-quality annotated data, and performing a more in-depth analysis of classifier subtype distinctions.



\bibliography{main}

\appendix

\section{Experiment Settings}

\subsection{Hardware}
All experiments were conducted using a single NVIDIA A100 GPU with 40GB of dedicated memory and a single NVIDIA H20 GPU with 96GB of memory, hosted on a system equipped with an AMD EPYC 9K84 96‑core processor (16 vCPUs) and 150 GB of system RAM. Each experimental run was configured to have a duration exceeding three hours.

\subsection{Experiement Setup}
This study involves Transformers packages with version 4.55.2 (Hugging Face) and platform of PyTorch with version 2.8.0.

\subsection{Prompts Usage}
We utilized the the prompt shown in table \ref{tab:Prompt-set} to make our model inference and fine tune. Due to the task is Chinese classifier prediction, we directly apply Chinese as the target language for prompt, the English version is also attached to the table for reference.

\begin{table*}[htbp]
\centering
\fontsize{10pt}{4pt}
\renewcommand{\arraystretch}{1.2} 
\begin{tabular}{@{}m{3.5cm}>{\raggedright\arraybackslash}m{12cm}@{}}
\toprule
\textbf{Prompt Type} & \textbf{Prompt Text} \\
\midrule
English & As a professional native speaker of Chinese, please complete the task of filling in the missing measure words. A Chinese sentence lacking a measure word will be input. Please identify and select only one single-character measure word that best fits the position indicated by the underscore “\_” in the sentence, according to the rules of Chinese measure word usage. Each sentence contains only one underscore. Please note that the final output should only include the measure word you selected, without any additional information.

Sentence lacking a measure word:

Output measure word: \\

\addlinespace[2pt]
Chinese & 你作为一个专业的中文母语者，现请你完成补全量词的任务。现在会输入一个缺乏量词的中文句子，请根据中文量词搭配规范，在输入的缺乏量词的句子中，在短下划线“\_”的位置，找出且仅找出一个最合适这个位置的单字量词，每一个句子只存在一个下划线，请注意最后的结果只输出你选择的量词，而没有其他的任何信息。

缺乏量词的句子：

输出的量词： \\

\bottomrule
\end{tabular}
\caption{The demonstration of prompts used in both the inference and fine-tuning stages is provided below. The English version is a direct, literal translation of the primarily used Chinese version.}
\label{tab:Prompt-set}
\end{table*}

\section{Instruction and Statistics of Classifier Annotation}
For the sake of systematic investigation of Chinese measure words in the experiment, the quantifiers are sorted into six classes: individual classifier, event classifier, kind classifier, container classifier, standard measure word, and approximate measure word. It was hand-annotated by 3 native Chinese speakers with research background in Chinese linguistics.

To assess the reliability and the objectivity of this annotation, we calculated the Inter-Annotator Agreement (IAA) through the application of Cohen’s kappa to a randomly selected sample of 500 classifier tokens. We thereby reached the IAA score of 82.68\%, the existence of which indicates high annotator agreement and the robustness of our scheme of classifier categorization. For example, Event Classifier had been used in the counting of action or event occurrences. It does classify events and not physical objects. Therefore, the diction example like {"场", "次", "趟"} can be categorized into Event Classifier. Finally, we got the statistics about the the Count and Frequency for each classifier type, which can be seen in Table \ref{tab:6 types classifiers' count and frequency}. It presents the frequencies of the six classifier types on the test set in numbers. The extreme imbalance of frequency often (e.g., Individual classifiers as the majority, Approximate/Standard classifiers as the minority) offers invaluable background in understanding the fine-grained model performance investigation by classifier type. It indicates the degree of difficulty and availability of data for each of the classifier types during the evaluation process.

\begin{table}[htbp]
\centering
\label{tab:6 types classifiers' count and frequency}
\begin{tabular}{lccc}
\toprule
Classifier Type & Count & Frequency (\%) \\
\midrule
Individual classifier & 1173 &  62.79\\
Kind Classifier & 247 & 13.22 \\
Event Classifier & 180 &  9.64\\
Container Measure & 134 & 7.17 \\
Standard & 57 &  3.05\\
Approximate Measure & 77 & 4.12 \\
\bottomrule
\end{tabular}
\caption{The count and frequency of 6 different types of classifiers in test set}
\label{tab:6 types classifiers' count and frequency}

\end{table}

\section{Chinese Classifier Categories and Explanation}
\label{ref: Chinese Classifier Categories and Explanation}
In Chinese, classifiers are a crucial grammatical category used in noun phrases to mark noun classes or quantify nouns. They can be broadly divided into two main types: sortal classifiers and measure words.

Sortal classifiers primarily serve to highlight inherent characteristics of the head noun, categorizing it based on shape, function, or other salient properties. They are further classified into three subtypes:

Individual classifiers classify concrete or abstract objects based on their natural units, such as 只 for animals (e.g., 一只猫 "a cat") or 张 for flat objects (e.g., 一张纸 "a piece of paper").

Event classifiers enumerate occurrences of events, like 次 for instances of actions (e.g., 一次会议 "one meeting") or 场 for performances (e.g., 一场比赛 "a match").

Kind classifiers categorize nouns by type rather than individual instances, such as 种 for kinds (e.g., 三种动物 "three types of animals").

On the other hand, measure words focus on quantifying the noun rather than classifying its inherent features. They include:

Container measure words, which denote quantity based on containers (e.g., 碗 "bowl" in 一碗饭 "a bowl of rice").

Standard measure words, which use fixed units of measurement (e.g.,米 "meter" in 三米长 "three meters long").

Approximation measure words, which indicate vague quantities (e.g.,些 "some" in 一些问题 "some problems").

\section{Relationship of classifier frequency and accuracy}
\label{ref: Relationship of classifier frequency and accuracy}

\begin{table}[htbp]  
\centering
\fontsize{10pt}{4pt}
\label{tab:classifiers-samples-per-group}
\begin{tabular}{@{}lcccc@{}}
\toprule
Group & NoC & Count & Proportion (\%) \\
\midrule
Group 1 & 1 & 677 & 34.95 \\
Group 2 & 1 & 113 & 5.83 \\
Group 3 & 1 & 102 & 5.27 \\
Group 4 & 4 & 120 & 6.20 \\
Group 5 & 10 & 116 & 5.99 \\
Group 6 & 15 & 120 & 6.20 \\
Group 7 & 15 & 120 & 6.20 \\
Group 8 & 15 & 120 & 6.20 \\
Group 9 & 15 & 120 & 6.20 \\
Group 10 & 21 & 118 & 6.09 \\
Group 11 & 40 & 120 & 6.20 \\
Group 12 & 34 & 91 & 4.70 \\
\bottomrule
\end{tabular}
\caption{This table demonstrates the rank based on the frequency of classifiers from the highest frequency of group 1 to lowest frequency group 12 and distribution of the classifiers. The NoC(number of categories) demonstrate how many categories of classifiers in this group.}
\label{tab:sample counts} 
\end{table}

\begin{figure}[h]
\centering
\includegraphics[width=0.48\textwidth]{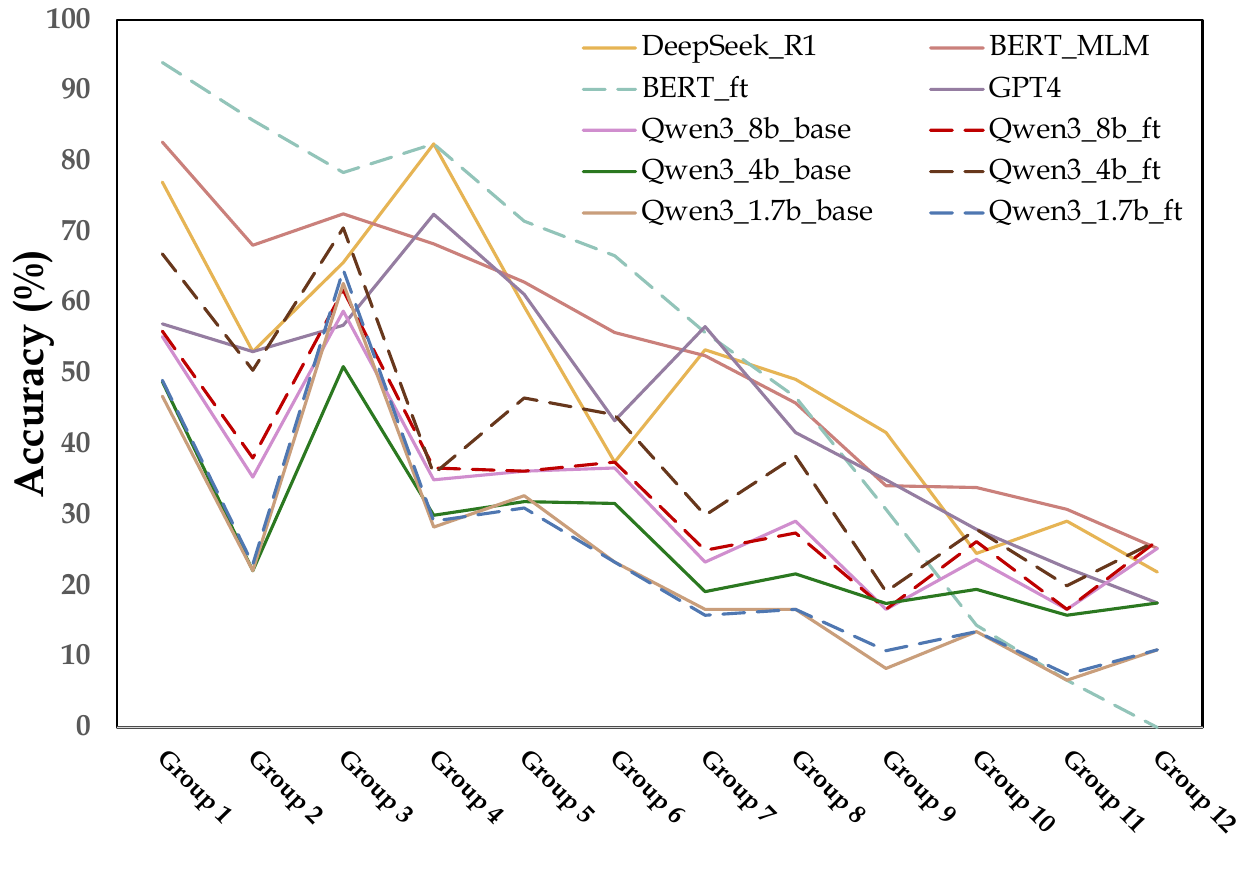}
\caption{Line chart of accuracy for used models in groups divided by classifier frequency from high to low. The dashed line represents the fine-tuned model, and the solid line represents the original model.} 
\label{fig:fig4}
\end{figure}

To explore the relationship between classifiers' frequency and accuracy, we sorted and grouped the classifiers that appeared in Chinese Classifier Dataset \cite{peinelt-etal-2017-classifierguesser} based on their frequency of occurrence. With the grouping results summarized in Table \ref{tab:sample counts}, we first assign the classifier exhibiting the highest frequency to Group 1. The classifiers with the second and third highest frequencies are then placed into Group 2 and Group 3, respectively. To achieve a more balanced distribution across categories, and considering that Groups 2 and 3 are of appropriate scale to accommodate additional classifiers, we establish a count threshold of 120 for group assignment. Any remaining classifiers that meet this threshold are sequentially allocated to subsequent groups in the processing order. We then separately aggregated the results of all applied models on the test set according to these groupings, as shown in Figure ~\ref{fig:fig4}. It can be observed that, overall, the accuracy of language models in predicting quantifiers decreases as the frequency of quantifiers in the dataset declines, indicating the prediction of classifier (or LogProb ranking) are influenced by the word frequency.


\end{CJK}
\end{document}